\documentclass[preprint,amsmath,amssymb]{revtex4-2}
\usepackage{mathtools}
\usepackage{bm}
\usepackage{braket}
\renewcommand{\boldsymbol}{\bm}
\DeclareMathOperator{\diag}{diag}
\DeclareMathOperator{\E}{\mathbb{E}}
\DeclareMathOperator{\Var}{Var}
\newcommand{\iid}{\overset{\text{iid}}{\sim}}
\renewcommand{\figurename}{Fig.}
\usepackage{graphicx}
\begin{document}
\title{Initialization Method for Factorization Machine Based on Low-Rank Approximation for Constructing a Corrected Approximate Ising Model}

\author{Yuya Seki}
\affiliation{Graduate School of Science and Technology, Keio University, Kanagawa 223-8522, Japan}

\author{Hyakka Nakada}
\affiliation{Graduate School of Science and Technology, Keio University, Kanagawa 223-8522, Japan}
\affiliation{Recruit Co., Ltd., Tokyo 100-6640, Japan}

\author{Shu Tanaka}
\affiliation{Graduate School of Science and Technology, Keio University, Kanagawa 223-8522, Japan}
\affiliation{Department of Applied Physics and Physico-Informatics, Keio University, Kanagawa 223-8522, Japan}
\affiliation{Keio University Sustainable Quantum Artificial Intelligence Center (KSQAIC), Keio University, Tokyo 108-8345, Japan}
\affiliation{Human Biology-Microbiome-Quantum Research Center (WPI-Bio2Q), Keio University, Tokyo 108-8345, Japan}

\begin{abstract}
This paper presents an initialization method that can approximate a given approximate Ising model with a high degree of accuracy using a factorization machine (FM), a machine learning model. The construction of an Ising models using an FM is applied to black-box combinatorial optimization problems using factorization machine with quantum annealing (FMQA). It is anticipated that the optimization performance of FMQA will be enhanced through an implementation of the warm-start method. Nevertheless, the optimal initialization method for leveraging the warm-start approach in FMQA remains undetermined. Consequently, the present study compares initialization methods based on random initialization and low-rank approximation, and then identifies a suitable one for use with warm-start in FMQA through numerical experiments. Furthermore, the properties of the initialization method by the low-rank approximation for the FM are analyzed using random matrix theory, demonstrating that the approximation accuracy of the proposed method is not significantly influenced by the specific Ising model under consideration. The findings of this study will facilitate advancements of research in the field of black-box combinatorial optimization through the use of Ising machines.
\end{abstract}

\maketitle

\section{Introduction}

Quantum annealing (QA)~\cite{kadowaki1998quantum,tanaka2017quantum,chakrabarti2023quantum} and simulated annealing (SA)~\cite{kirkpatrick1983optimization,johnson1989optimization,aragon1991optimization} are promising metaheuristics for combinatorial optimization problems.
Solving a combinatorial optimization problem is often extremely difficult due to an exponentially large search space and complex shape of the objective function.
To address the difficulty, QA (SA) exploits quantum (temperature) effects to search for the optimal solution to the combinatorial optimization problem.
The development of annealing devices that implement QA or SA has enabled the rapid solution of combinatorial optimization problems~\cite{Johnson2011quantum,Barends2016digitized,inagaki2016coherent,Rosenberg20173dintegrated,Maezawa2019toward,Novikov2018exploring,Mukai2019superconducting,Tsukamoto2017accelerator,Aramon2019,Yamaoka2016,Goto2019combinatorial,mohseni2022ising}. For example, a device that performs quantum annealing has been developed, and a system comprising over 5,000 superconducting qubits is currently available for use~\cite{king2023quantum}. Moreover, the development of annealing devices that do not utilize quantum effects is also currently underway.
These annealing devices are collectively referred to as Ising machines.

To use Ising machines, it is necessary to express the objective function of a combinatorial optimization problem as quadratic unconstrained binary optimization (QUBO) form or as a Hamiltonian of the Ising model that is equivalent to the QUBO form. It is well-established that objective functions of numerous combinatorial optimization problems can be expressed as Hamiltonians of the Ising model~\cite{lucas2014ising}.
However, there are cases where the objective function cannot be expressed in an exact form using the Ising model.
In particular, it is not feasible to derive a Hamiltonian of the Ising model for black-box combinatorial optimization problems.
This is due to the fact that, in the case of black-box combinatorial optimization problems, the analytical form of the objective function, referred to as the ``black-box function,'' is not provided, and thus it is not feasible to formulate the objective function using an Ising model.

Factorization machine with quantum annealing (FMQA)~\cite{kitai2020designing}, which is a model-based optimization method, is a cutting-edge solution to the above issue.
In FMQA, the black-box function is modeled by an Ising model so that input-output relationships associated with the black-box function are reproduced by the Ising model.
Here, the modeling is performed by a machine learning model called factorization machine (FM)~\cite{rendle2010factorization}.
Subsequently, the Ising machine is employed to identify the optimal solution of the Ising model.
If the Ising model accurately represents the black-box function, we expect to find an optimal solution to the black-box combinatorial optimization problem.
The training dataset to improve the accuracy is constructed from samples explored by the Ising machine.
By alternating between sampling with the Ising machine and training the FM, the optimal solution is sought while simultaneously generating the samples necessary for training.
Since the modeling process must be performed several times, the time required for this process can become a limiting factor of FMQA.
FMQA attains high-speed calculations through the utilization of FM, which possesses a minimal number of model parameters.
Moreover, FMQA necessitates the estimation of an Ising model from a limited number of samples.
According to Ref.~\cite{rendle2010factorization}, since the FM accounts for the correlation between the components of training data points, it is possible to estimate a highly accurate model even from a small number of samples.
As can be observed, there are numerous advantages associated with the utilization of the FM for the estimation of Ising models.
The advent of FMQA has enabled the usage of Ising-machine-based high-speed solution searches for problems that were previously considered incompatible with Ising machines. 
This has led to numerous application studies~\cite{kitai2020designing,inoue2022towards,seki2022black,matsumori2022application,gao2023quantum,nawa2023quantum,tucs2023quantum,okada2023design,couzinie2024machine,hida2024topology,
xiao2024application}.

A novel application of FMQA is enhancement of the precision of approximate Ising models.
Examples of applied research on Ising machines that derive an approximate Hamiltonian of the Ising model and perform solution searches can be found in~\cite{nagel2020up,sampei2023quantum,endo2022phasefield}.
Although the precise objective functions for these problems are black-box functions, approximate analytical objective functions are provided as the Ising model.
An issue with these applications is that if the approximation is of insufficient accuracy, it becomes challenging to search for better solutions to the precise objective function.
Recently a method for addressing this issue has been proposed~\cite{aoki2025formulation}.
This method uses the FM to derive a correction term for the approximate Hamiltonian of the Ising model, resulting in better solution to the precise objective function.
Another potential solution to the issue is to employ FMQA in the following manner.
The first step is to express the approximate Hamiltonian of the Ising model in the form of the FM.
Subsequently, the FM model is refined to represent the precise objective function through training.
By employing this refined Ising model in solution searches, it is expected that better solutions can be identified.
This approach can be classified as a warm-start method in machine learning and mathematical optimization.
To apply the warm-start method to FMQA, it is necessary to develop an initialization method capable of reproducing the approximate Hamiltonian of the Ising model using the FM.

This paper focuses on the first step of the warm-start method.
The aim of this paper is to shed light on the initialization method of the FM that reproduces an approximate Hamiltonian of the Ising model with  greater accuracy.
To this end, we investigate two types of initialization methods: initialization with a low-rank approximation and random initialization with a probability distribution that is appropriately constructed.
Our findings, based on numerical experiments and analysis using random matrix theory, demonstrate that the initialization method with a low-rank approximation can reproduce the approximate Ising model with greater accuracy than the random initialization.

The subsequent sections of this paper are organized as follows.
In Sec.~\ref{sec:factorization_machine}, we introduce the FM, which is the machine learning model utilized in this study.
Subsequently, Sec.~\ref{sec:initialization_methods} delineates the initialization methods for the FM.
In Sec.~\ref{sec:approximation_error_analysis_random_matrix_theory}, we introduce an analysis method based on random matrix theory. In Sec.~\ref{sec:results}, we present the results of numerical experiments and analysis based on random matrix theory to determine the initialization method for the FM with the desired properties. Finally, in Sec.~\ref{sec:conclusion}, we provide a summary of the findings.

\section{Factorization Machine}
\label{sec:factorization_machine}

Factorization machines (FMs) constitute a specific type of machine learning model that can be utilized for regression, classification, and ranking problems~\cite{rendle2010factorization}.
The FM model is applied to a real vector of dimension $N$, $\bm{x} = (x_{1}, \dotsc ,x_{N}) \in \mathbb{R}^{N}$, and uses the following model equation, which returns a real number:
\begin{align}\label{eq:fm_model_equation}
f_{+}(\boldsymbol{x}; \boldsymbol{\theta}) = w_{0} + \sum_{i=1}^{N}w_{i}x_{i} + \sum_{\substack{i,j=1\\(i<j)}}^{N} \left< \boldsymbol{v}_{i}, \boldsymbol{v}_{j} \right> x_{i}x_{j},
\end{align}
where $w_{0} \in \mathbb{R}$, $w_{i} \in  \mathbb{R}$, $\boldsymbol{v}_{i} \in \mathbb{R}^{K}$ are the model parameters of the FM determined by learning training data, and these are collectively represented as $\boldsymbol{\theta}$.
Here, the natural number $K$ is a hyperparameter of the FM.
The sum of the quadratic terms is taken for all pairs of integers $i$ and $j$ satisfying $i<j$, where $i$ and $j$ are two integers taking values from $1$ to $N$.
In addition, the coupling coefficient of the quadratic term, $\left< \boldsymbol{v}_i, \boldsymbol{v}_j\right>$, represents the inner product of the two vectors $\boldsymbol{v}_i$ and $\boldsymbol{v}_j$.
In this paper, the matrix whose $(i,j)$ component is $\langle \boldsymbol{v}_i, \boldsymbol{v}_j\rangle$ is referred to as the coupling matrix of the FM.
In this case, the hyperparameter $K$ represents the upper bound for the rank of the coupling matrix of the FM.
For simplicity, henceforth, we will express the hyperparameter $K$ as the rank of the FM.

The coupling matrix of the FM in the form of inner products offers two advantages for FMQA.
First, as stated in Ref.~\cite{rendle2010factorization}, the coupling coefficient based on the inner product incorporates correlations between the components of the input variable.
This property enables the estimation of a highly accurate model even with a limited amount of training data.
Second, it reduces the amount of computation required.
Although the FM includes the quadratic terms for the input variables, it is established that the time complexity for forward calculations is $O(KN)$.
Moreover, the time complexity for calculating the partial derivatives necessary for estimating the model parameters during training is also $O(KN)$.
The time-efficient training of the FM plays a significant role in reducing the calculation time of FMQA.
Since FMQA requires repetitive training of the FM model within a single procedure, training is a possible bottleneck.
The linear time complexity of training ensures that the computational time required for training is not likely to be a significant issue.

However, there is a limitation to the representational ability of the FM, which is constrained by the fact that the coupling coefficients are expressed as inner products. It is not feasible to express arbitrary coupling coefficients in the form of inner products unless a sufficiently large $K$ is employed (see Sec.~\ref{sec:init_with_eigen_decomposition} for details). 
Consequently, even if a true model that generates the training data is a quadratic function of the input variables, it is generally not possible to completely express that training data using the FM.

A true model that can be expressed by an FM depends on the sign in the model equation.
This property is referred to as the asymmetry of FMs~\cite{prillo2017anelementary}.
As a consequence of this asymmetry, the behavior of the positive FM, as given by Eq.~\eqref{eq:fm_model_equation}, and the negative FM, as given below, are distinct:
\begin{align}
\label{eq:negative_fm}
f_{-}(\boldsymbol{x}; \boldsymbol{\theta}) = -w_{0} - \sum_{i=1}^{N}w_{i}x_{i} - \sum_{\substack{i,j=1\\(i<j)}}^{N}\left< \boldsymbol{v}_{i},\boldsymbol{v}_{j} \right>x_{i}x_{j}.
\end{align}
Indeed, it has been demonstrated that the learning performance differs between the positive FM and the negative FM~\cite{lin2019factorization}. 
In this study, numerical experiments are conducted for both the positive and the negative FM, and the model exhibiting superior performance is selected.

This paper addresses the regression problem through the use of the FM with input components that assume a value of either $+1$ or $-1$.
In the regression problem, the objective is to identify the model parameters of the FM, $\boldsymbol{\theta}$, so as to minimize a loss function for given training data, $\mathcal{D} = \set{ (\boldsymbol{x}_{d}, y_{d}) \in \set{ +1, -1 }^{N} \times \mathbb{R} | d=1,\dotsc , D }$. In this paper, the mean-squared error, as defined below, is employed as the loss function:
\begin{align}\label{eq:loss_function}
L(\boldsymbol{\theta}) = \frac{1}{D}\sum_{d=1}^{D}\left( f(\boldsymbol{x}_{d}; \boldsymbol{\theta}) - y_{d} \right)^{2},
\end{align}
where the function $f$ is either $f_{+}$ or $f_{-}$.
In training, the model parameters, $\boldsymbol{\theta}$, are updated repeatedly in the direction that minimizes the loss function represented by Eq.~\eqref{eq:loss_function}.
The number of updates required until the loss function converges depends on the initial value of the model parameters.
Therefore, to reduce the calculation time required for training, it is necessary to devise a method for initializing the FM.

\section{Initialization Methods for Factorization Machine}
\label{sec:initialization_methods}

In this section, we introduce the method for initializing the model parameters of the FM.
In this study, we assume that an approximate Ising model, which is a true model for FM, is available. The Hamiltonian of $N$ Ising variables, $\boldsymbol{x} = (x_{1}, \dotsc ,x_{N}) \in \set{ +1, -1}^{N}$, is represented by
\begin{align}
H(\boldsymbol{x}) &= c - \sum_{i=1}^{N}h_{i}x_{i}  -\frac{1}{2}\sum_{\substack{i,j=1\\(i \ne j)}}^{N}J_{ij}x_{i}x_{j}\notag\\
\label{eq:apporx_ising_model}
&= c - \sum_{i=1}^{N}h_{i}x_{i}  -\sum_{\substack{i,j=1\\(i < j)}}^{N}J_{ij}x_{i}x_{j},
\end{align}
where $c \in \mathbb{R}$ is a constant, $h_{i} \in \mathbb{R}$ is a local magnetic field coefficient, and $J_{ij} \in \mathbb{R}$ is a coupling coefficient.
Here, the coupling coefficient is symmetric with respect to the subscript: $J_{ij}=J_{ji}$.
For example, it is established that the cost function to be minimized in the problem of minimizing the energy of molecular adsorption on a catalyst surface~\cite{sampei2023quantum} and the problem of model compression in machine learning~\cite{nagel2020up} can be approximated by an Ising model.
In order to enhance the precision of these specified approximate Ising models by the warm-start FMQA, as described in the introduction, it is essential to employ a methodology for initialization of the FM model that is in close proximity to the approximate Ising model.

The following sections introduce two FM model initialization methods.
The first is a method that generates an initial FM model through a low-rank approximation method of a coupling matrix $J=(J_{ij})$ by the eigenvalue decomposition.
The second method generates an initial FM model that is close to the given approximate Ising model by adjusting a probability distribution that the model parameters follow.

\subsection{Initialization by Low-Rank Approximation of Coupling Matrix}
\label{sec:init_with_eigen_decomposition}

To generate an FM model that is close to an approximate Ising model, we only need to consider how to initialize the model parameters for the coupling matrix of the FM, $\boldsymbol{v}_i,\ (i=1,\dotsc, N)$.
This is because the constant and the linear terms of the FM model can be determined so that they match those of the given approximate Ising model:
\begin{align}
\begin{cases}
    \begin{aligned}
    w_0 &= +c,\\
    w_i &= -h_i\ (i=1,\dotsc , N)     
    \end{aligned}
    & \text{for positive FM},\\[10pt]
    \begin{aligned}
    w_0 &= -c,\\
    w_i &= +h_i\ (i=1,\dotsc , N)     
    \end{aligned}
    & \text{for negative FM}.
\end{cases}
\end{align}
Accordingly, we propose a methodology for approximating the coupling coefficient of the FM to that of the approximate Ising model:
\begin{align}
\left< \boldsymbol{v}_{i}, \boldsymbol{v}_{j} \right> \approx
\begin{cases}
- J_{ij} & \text{for positive FM}, \\
+ J_{ij} & \text{for negative FM}.
\end{cases}
\end{align}
In the following, we will limit our consideration to the negative FM without loss of generality. The method for approximating the positive FM is obtained by reversing the sign of the coupling coefficient $J_{ij}$.

The parameters of the quadratic terms of the FM can be determined by the low-rank approximation of the coupling matrix, denoted by $J=(J_{ij})$. In this case, low-rank approximation is possible using the eigenvalue decomposition.
Since the coupling matrix $J$ is real symmetric, it is diagonalizable:
\begin{align}
J = U \Sigma U^\top.
\end{align}
Here, the matrix $U$ is a real orthogonal matrix comprising the eigenvectors of $J$, and $\Sigma$ is a diagonal matrix comprising the eigenvalues $\lambda_{1} \ge \lambda_{2} \ge \dotsb \ge \lambda_{N}$ as its components.
Here, the eigenvalues of the coupling matrix $J$ are shifted by the smallest eigenvalue $\lambda_{N}$ so that all eigenvalues are non-negative:
\begin{align}\label{eq:shifted_coupling_matrix}
J' &= J - \lambda_{N} I,
\end{align}
where $I$ is the $N$-dimensional identity matrix.
It should be noted that the value of the Hamiltonian remains unaltered even when the coupling matrix in Eq.~\eqref{eq:apporx_ising_model} is replaced with the shifted matrix $J'$.
Furthermore, the matrix $J'$ can be diagonalized using the same matrix $U$ as the coupling matrix $J$:
\begin{align}\label{eq:decomposition_of_J}
J' = U \Sigma' U^\top.
\end{align}
Here, we have
\begin{align}\label{eq:shifted_eigenvalues_matrix}
\Sigma' = \begin{pmatrix}
\lambda_{1}' &&& \\
& \ddots && \\
&& \lambda_{N-1}'  & \\
&&& 0
\end{pmatrix},
\end{align}
where $\lambda'_{i} = \lambda_{i} - \lambda_{N}$.
By using the diagonal matrix formed by taking the $K$ largest eigenvalues
\begin{align}
\Sigma'_{K} = \begin{pmatrix}
\lambda'_{1} && \\
& \ddots & \\
&& \lambda_{K}'
\end{pmatrix},
\end{align}
and $N \times K$ matrix $U_{K}$, which has the corresponding eigenvectors in its columns, the coupling matrix is approximated as
\begin{align}\label{eq:low_rank_approximation_of_J}
J' \approx U_{K}\Sigma'_{K}U_{K}^\top.
\end{align}
We will discuss this approximation accuracy later.
Since the diagonal matrix $\Sigma'_{K}$ is semi-positive definite, its square root can be defined. Thus, by writing
\begin{align}
(\boldsymbol{v}_{1}, \dotsc , \boldsymbol{v}_{N}) = (\Sigma'_{K})^{1/2}U_{K}^\top,
\end{align}
the relation $\left< \boldsymbol{v}_{i}, \boldsymbol{v}_{j} \right> \approx  J_{ij}$ is satisfied.
Since the smallest eigenvalue of the coupling matrix, $J'$ is $0$, it follows that any coupling matrix can be exactly expressed in exact form by the FM, provided that $K=N-1$.

The approximation accuracy of the low-rank approximation of the expression~\eqref{eq:low_rank_approximation_of_J} is evaluated as follows.
The Frobenius norm of the difference between the shifted coupling matrix, denoted by $J'$, and the approximate coupling matrix is employed as an indicator of the approximation accuracy.
In this case, the following formula is obtained:
\begin{align}
\delta J'_{k} &= \lVert J' - U_{k}\Sigma'_{k}U_{k}^\top \rVert \notag\\
\label{eq:theoretical_coupling_error_low_rank_approx}
&= \sqrt{\sum_{i=k+1}^{N}(\lambda_{i}')^{2}}.
\end{align}
In other words, the smaller the sum of the squares of the ignored eigenvalues, the smaller the error due to the low-rank approximation.
This indicates that the low-rank approximation is more feasible when the eigenvalue distribution of the coupling matrix is biased towards small values. Furthermore, the FM hyperparameter $K$ can be determined in a way that ensures the value of Eq.~\eqref{eq:theoretical_coupling_error_low_rank_approx} is less than a certain allowable value.

\subsection{Random Initialization}
\label{sec:random_initialization_methods}
As random initialization methods, this study introduces an initialization method based on the energy distribution and that based on the coupling coefficient distribution.
Similar to the initialization method based on the low-rank approximation in the previous section~\ref{sec:init_with_eigen_decomposition}, this method focuses only on the quadratic terms.

\subsubsection{Random Initialization Based on Energy Distribution}

In the initialization method based on energy distribution, the model parameters are initialized so that the distribution of the interaction energy of the approximate Ising model and that of the FM model are aligned.
The mean and variance are used as the features that characterize the energy distribution.
As the distribution that each component of the input variable $\boldsymbol{x}$ follows is unknown, we consider the energy distribution when each component is determined completely at random.
This is equivalent to considering the energy distribution in the high-temperature limit.

First, we calculate the mean and variance of the interaction energy of an approximate Ising model in Eq.~\eqref{eq:apporx_ising_model} in the high-temperature limit.
The interaction energy is given by
\begin{align}
H_{2}(\boldsymbol{x}) = - \sum_{\substack{i,j=1\\(i<j)}}^{N}J_{ij}x_{i}x_{j}
\end{align}
At sufficiently high temperatures, the input variables take on the values $+1$ and $-1$ independently and with equal probability:
\begin{align}
\E_{\boldsymbol{x}} [x_{i}] =0,\quad \E_{\boldsymbol{x}} [x_{i}^{2}] =1 \quad \text{for all $i$}.
\end{align}
Here, $\E_{\boldsymbol{x}}$ represents the average with regard to the distribution of the input variable $\boldsymbol{x}$.
First, the average of $H_{2}(\boldsymbol{x})$ is given by
\begin{align}
\E_{\boldsymbol{x}} [H_{2}(\boldsymbol{x})] = -\sum_{\substack{i,j=1\\(i<j)}}^{N}J_{ij}\E_{\boldsymbol{x}}[x_{i}] \E_{\boldsymbol{x}} [x_{j}] = 0.
\end{align}
The equality in the above equation holds regardless of the specific value of $J_{ij}$.
Next, the variance of $H_{2}$ can be written as follows:
\begin{align}
  \Var_{\boldsymbol{x}}[H_{2}(\boldsymbol{x})] &= \sum_{\substack{i,j=1\notag\\(i<j)}}^{N}\sum_{\substack{l,m=1\\(l<m)}}^{N}J_{ij}J_{lm}\E_{\boldsymbol{x}}[x_{i}x_{j}x_{l}x_{m}]\\
  \label{eq:var_ising_coupling_term}
  &= \sum_{\substack{i,j=1\\(i<j)}}^{N}J_{ij}^{2}.
\end{align}
Here, $\Var_{\boldsymbol{x}}$ denotes the variance with regard to the distribution of $\boldsymbol{x}$.
In Eq.~\eqref{eq:var_ising_coupling_term}, we used the fact that input variables with different indices follow independent probability distributions.
Since the variance of the expression~\eqref{eq:var_ising_coupling_term} differs for each instance, we use the instance average as a feature for the variance of the interaction energy:
\begin{align}\label{eq:instance_avg_ising_coupling_energy_var}
\E_{\boldsymbol{J}}\left[ \Var_{\boldsymbol{x}}[H_{2}(\boldsymbol{x})] \right] &= \frac{N(N-1)}{2}\E_{\boldsymbol{J}}[J_{ij}^{2}],
\end{align}
where $\E_{\boldsymbol{J}}$ refers to the average with regard to a distribution of coupling coefficients which is determined by the approximate Ising model.

Next, we calculate the mean and variance of the interaction energy of the positive FM and the negative FM at high temperatures.
The interaction energy is given by
\begin{align}\label{eq:fm_second_term}
f_{2,\pm}(\boldsymbol{x}) = \pm \sum_{\substack{i,j=1\\(i<j)}}^{N}\left< \boldsymbol{v}_{i}, \boldsymbol{v}_{j} \right>x_{i}x_{j},
\end{align}
Here, we assume that the model parameters of the coupling coefficients of the FM, $\boldsymbol{v}_{i}=(v_{i1}, ..., v_{iK})$, $(i=1, ..., N)$, follow the following independent Gaussian distribution for each component $v_{ik}$:
\begin{align}\label{eq:distribution_v_energy_init}
v_{ik} \iid \mathcal{N} \left( \mu_{\mathrm{v}}, \sigma_{\mathrm{v}}^{2} \right).
\end{align}
The average value of Eq.~\eqref{eq:fm_second_term} with regard to $\boldsymbol{x}$ is $0$, as in the case of the approximate Ising model.
The variance of Eq.~\eqref{eq:fm_second_term} can be written as follows:
\begin{align}
  \Var_{\boldsymbol{x}}[f_{2,\pm}(\boldsymbol{x})]
  &= \sum_{\substack{i,j=1\\(i<j)}}^{N}\sum_{\substack{l,m=1\\(l<m)}}^{N}\sum_{k,k'=1}^{K}v_{ik}v_{jk}v_{lk'}v_{mk'}\E_{\boldsymbol{x}} [x_{i}x_{j}x_{l}x_{m} ] \notag \\
  \label{eq:var_fm_second_term}
  &=\sum_{\substack{i,j=1\\(i<j)}}^{N}\sum_{k,k'=1}^{K}v_{ik}v_{jk}v_{ik'}v_{jk'}.
\end{align}
Since this variance depends on the model parameters, we use the mean value of the model parameter distribution~\eqref{eq:distribution_v_energy_init} as a feature:
\begin{align}
\E_{\boldsymbol{v}}\left[\Var_{\boldsymbol{x}}[f_{2,\pm}(\boldsymbol{x})]\right]
\label{eq:expected_fm_second_energy_var}
&= \frac{KN(N-1)}{2}\left(\sigma_{\mathrm{v}}^4 + 2\mu_{\mathrm{v}}^2\sigma_{\mathrm{v}}^2 +K\mu_{\mathrm{v}}^4\right).
\end{align}
Here, $\E_{\boldsymbol{v}}$ denotes the average with regard to the distribution in Eq.~\eqref{eq:distribution_v_energy_init}.
To obtain Eq.~\eqref{eq:expected_fm_second_energy_var}, we used the independence of the random variables $v_{ik}$.

We derive the conditions for the interaction energy distribution of the approximate Ising model to approach that of the FM model.
First, the mean values of the distributions are both $0$, and thus they are the same.
Next, in order for the variances of the distributions to approach each other, it is sufficient for the following two equations~\eqref{eq:instance_avg_ising_coupling_energy_var} and~\eqref{eq:expected_fm_second_energy_var} to be equal.
Since it is not possible to determine both $\mu_{\mathrm{v}}$ and $\sigma_{\mathrm{v}}$ simultaneously under the condition, we have set $\mu_{\mathrm{v}}=0$.
Hence, in order to initialize an FM with an energy distribution that is as close as possible to that of the approximate Ising model, it is sufficient to generate the model parameters for the FM coupling coefficients from the following probability distribution:
\begin{align}\label{eq:energy_based_init}
v_{ik} \iid \mathcal{N}\left( 0, \sqrt{\frac{\E_{\boldsymbol{J}}[J_{ij}^{2}]}{K}} \right).
\end{align}
Since the coupling coefficients of the approximate Ising model are given, the average on the right-hand side of Eq.~\eqref{eq:energy_based_init} can be replaced with the following sample average:
\begin{align}
\E_{\boldsymbol{J}}[J_{ij}^{2}] = \frac{2}{N(N-1)}\sum_{\substack{i,j=1\\(i<j)}}^{N}J_{ij}^{2}.
\end{align}

\subsubsection{Random Initialization Based on Coupling Constants Distribution}

In what follows, we derive a condition under which features of the distribution of coupling coefficients of the FM agree well with those of an approximate Ising model.
As with the previous analysis for energy distribution, the mean and variance of the distribution of coupling coefficient are employed as features.
As will be demonstrated subsequently, if the mean of the distribution of $J_{ij}$ is non-negative, the mean and variance of the distribution can be made to align with those of the coupling coefficient distribution of the negative FM.
In the case of a negative mean of the distribution of $J_{ij}$, it is shown that the positive FM can be used to align the distribution of coupling coefficient of the FM with that $J_{ij}$ in a similar manner.
Accordingly, in the following, we will limit our consideration to the case where the mean of the distribution of $J_{ij}$ is non-negative.

Let the mean of the coupling coefficient $J_{ij}$ be $\mu \ge 0$ and the variance be $\sigma^{2}$.
The negative FM model parameter $v_{ik}$ is assumed to be an independent random variable that follows the Gaussian distribution in Eq.~\eqref{eq:distribution_v_energy_init}.
Here, the condition so as to satisfy the following relations is considered:
\begin{align}
\label{eq:condition_mean_vivj_coupling_init}
\E_{\boldsymbol{v}} \left[ \left<\boldsymbol{v}_{i}, \boldsymbol{v}_{j} \right> \right] &= \mu, \\
\label{eq:condition_var_vivj_coupling_init}
\Var_{\boldsymbol{v}} \left[ \left<\boldsymbol{v}_{i}, \boldsymbol{v}_{j} \right> \right] &= \sigma^{2},
\end{align}
where $\E_{\boldsymbol{v}}$ and $\Var_{\boldsymbol{v}}$ represent the mean and variance with respect to the distribution in Eq.~\eqref{eq:distribution_v_energy_init}, respectively.
First, the left-hand side of Eq.~\eqref{eq:condition_mean_vivj_coupling_init} is written as
\begin{align}\label{eq:mean_vivj_coupling_init}
\E_{\boldsymbol{v}} \left[ \left<\boldsymbol{v}_{i}, \boldsymbol{v}_{j} \right> \right] &= K\mu_{\mathrm{v}}^{2}.
\end{align}
Note that since Eq.~\eqref{eq:mean_vivj_coupling_init} is always non-negative, it is not possible to match the mean of the distribution of coupling coefficients with the negative FM when $\mu < 0$.
Next, the left-hand side of Eq.~\eqref{eq:condition_var_vivj_coupling_init} is calculated as
\begin{align}\label{eq:var_vivj_coupling_init}
\Var_{\boldsymbol{v}} \left[ \left<\boldsymbol{v}_{i}, \boldsymbol{v}_{j} \right> \right] &= K\sigma_{\mathrm{v}}^{2}\left( \sigma_{\mathrm{v}}^{2}+2\mu_{\mathrm{v}}^{2} \right)
\end{align}
From Eqs.~\eqref{eq:condition_mean_vivj_coupling_init}--\eqref{eq:var_vivj_coupling_init}, it is sufficient to generate the model parameters with the negative FM from the following probability distribution:
\begin{align}
v_{ik} \iid \mathcal{N}\left( \sqrt{\frac{\mu}{K}}, \frac{\sqrt{\mu^{2}+K\sigma^{2}}-\mu }{K} \right).
\end{align}
Since the coupling coefficients of the approximate Ising model are given, $\mu$ is replaced with the sample mean and $\sigma^{2}$ is replaced with the unbiased variance.

\subsection{Reducing Error by Training}
This section presents a method to improve the precision of initialized FMs to the approximate Ising model further through model training.
First, the FM is initialized using one of the initialization methods described above for the given approximate Ising model.
Subsequently, the FM model is trained using training data $\mathcal{D}=\set{(\boldsymbol{x}_d, H(\boldsymbol{x}_d)) | d=1,\dotsc,D}$, where $H(\boldsymbol{x}_d)$ is the energy of the approximate Ising model given as Eq.~\eqref{eq:apporx_ising_model}.
The FM model that shows the highest precision during the training is selected as the initial model for the warm-start FMQA.
The model with the highest precision is defined as the one that minimizes an error measured by the Frobenius norm:
\begin{align}\label{eq:coupling_error}
\Delta J = \lVert J - \diag{J} - (G - \diag{G}) \rVert.
\end{align}
Here, $J$ is the coupling matrix of the approximate Ising model, and $G$ is the coupling matrix of FM.
The matrices $\diag{J}$ and $\diag{G}$ are diagonal matrices that have diagonal elements of $J$ and $G$, respectively.
The diagonal elements of $J$ and $G$ are subtracted since they do not appear in the Hamiltonian, and therefore do not impact the precision.
In this paper, the error defined in Eq.~\eqref{eq:coupling_error} is referred to as the coupling error.
The coupling error is a direct measure of precision of the FM, because the FM is identical with the approximate Ising model if and only if $\Delta J = 0$.

It should be noted that the coupling error is distinct from the error resulting from the low-rank approximation, as defined in Eq.~\eqref{eq:theoretical_coupling_error_low_rank_approx}.
Consider the initial coupling matrix of the FM by the low-rank approximation: $G=U_k\Sigma_k'U_k^\top$.
For the initial coupling matrix, the following inequality holds:
\begin{align}
\left(\delta J_k'\right)^2 &=\left\| J - \diag{J} - (G - \diag{G}) \right\|^2 + \left\| \diag{J} - \diag{G} - \lambda_N I\right\|^2 \notag \\
&\ge \left(\Delta J\right)^2. 
\end{align}
Consequently, the error $\delta J_k'$ provides an upper bound for the coupling error in the initialization method based on the low-rank approximation.

\section{Approximation Error Analysis with Random Matrix Theory}
\label{sec:approximation_error_analysis_random_matrix_theory}
In this study, we analyze the approximation accuracy of the initialization method based on the low-rank approximation, as introduced in Sec.~\ref{sec:init_with_eigen_decomposition}, using random matrix theory.
This analysis demonstrates that as the input dimension increases, the instance dependence of the approximation error in Eq.~\eqref{eq:theoretical_coupling_error_low_rank_approx} is eliminated.
Furthermore, we present a method for determining the effective rank $K^*$, which ensures sufficient approximation accuracy.

We consider an approximate Ising model where the components of the coupling matrix $J$ are distributed according to a Gaussian distribution:
\begin{align}
J_{ij} \overset{\textrm{iid}}{\sim} \mathcal{N}\left(\mu,\sigma^2\right), \label{eq:q_gauss}
\end{align}
where $i< j$, the mean of the coupling coefficient is denoted by $\mu$, the variance by $\sigma^{2}$, and the input dimension is $N$. In addition, we assume that diagonal $J_{ii}$ follows $\mathcal{N}\left(\mu,2\sigma^2\right)$. Thus, the coupling matrix $J$ is a Gaussian orthogonal random matrix.
In the case of a matrix with components that follow a Gaussian distribution with a mean of zero, the effective rank, $K^*$, can be readily derived, as the eigenvalue distribution can be obtained analytically using random matrix theory.
Nevertheless, this is not straightforward when dealing with a non-zero mean.
In this section, we will combine Wigner's semicircle law~\cite{wigner1955semicircle, wigner1958semicircle, tao2013semicircle} and the theory of the largest eigenvalue distribution theory~\cite{lang1964noncenterd,forrester2023noncenterd} to derive the effective rank of the coupling matrix $J$ in Eq.~\eqref{eq:q_gauss}.

First, we estimate the largest, the second largest, and the smallest eigenvalues ($\lambda_1, \lambda_2, \lambda_N$) of $J$ to approximate the eigenvalue distribution.
The accuracy of this approximation is confirmed in Sec.~\ref{sec:results}.
In the following analysis, it is assumed that the eigenvalues are arranged in descending order.
Using a random matrix $\bar{J}$ where each component follows a Gaussian distribution with a mean of zero and a variance $\sigma^2$, we obtain:
\begin{align}
J &= \mu\bm{1}\bm{1}^\top +\bar{J}.
\label{eq:q_tra}
\end{align}
Here, $\bm{1}$ is an $N$-dimensional all-ones column vector.
Since $\mu\bm{1}\bm{1}^\top$ is a real symmetric matrix, it can be diagonalized into $\Sigma_{\mu}$ using an appropriate real unitary matrix $U_{\mu} = (\bm{u}_1, \dotsc, \bm{u}_N)$.
The matrix $\mu\bm{1}\bm{1}^\top$ has an eigenvalue $N\mu$, and the remaining eigenvalues are zero.
Hence, the rank decreases to one as the variance of the coupling coefficient, $\sigma^2$, approaches zero, whereas it approaches full rank as the variance increases.
Assuming $\sigma$ is sufficiently smaller than $\sqrt{N}\mu$, we can expand the eigenvalues of the matrix in Eq.~\eqref{eq:q_tra} perturbatively:
\begin{align}
\lambda_i &\approx \lambda_{i}^{(0)}+\bm{u}_{i}^\top \bar{J} \bm{u}_{i}.
\label{eq:q_eigen}
\end{align}
Here, the eigenvalues of $\mu\bm{1}\bm{1}^\top$ are represented by $\lambda^{(0)}_i\ (i = 1, 2, \dotsc, N)$, i.e., $\lambda^{(0)}_1=N\mu, \lambda^{(0)}_2=\lambda^{(0)}_3=\cdots=\lambda^{(0)}_N=0$.
Thus, introducing finite means is considered to have the strongest effect on the largest eigenvalue.
In the following, we assume that the cumulative distribution of eigenvalues can be approximated using that of a random matrix with a mean of zero, except for $\lambda_1$.
According to Wigner's semicircle law~\cite{wigner1958semicircle}, the largest and smallest eigenvalues of $\bar{J}$ approach $2\sqrt{N}\sigma$ and $-2\sqrt{N}\sigma$, respectively, in the limit of $N\to\infty$.
If the system size $N$ is sufficiently large, the largest eigenvalue and second largest eigenvalue of $\bar{J}$ will converge asymptotically to the same value.
Furthermore, it is established that the largest eigenvalue $\lambda_1$ of the coupling matrix $J$ with a finite mean approaches $N\mu+\sigma^2/\mu$ in the limit of $N\to\infty$~\cite{lang1964noncenterd,forrester2023noncenterd}.
Thus, it follows that the largest eigenvalue $\lambda_1$, the second largest eigenvalue $\lambda_2$, and the smallest eigenvalue $\lambda_N$ of the coupling matrix $J$ approximately satisfy the following:
\begin{align}
\label{eq:eigen_1}
&\E\left[\lambda_1\right] \approx N\mu+\sigma^2/\mu,\\
\label{eq:eigen_2}
&\E\left[\lambda_2\right] \approx 2\sqrt{N}\sigma, \\
\label{eq:eigen_N}
&\E\left[\lambda_N\right] \approx -2\sqrt{N}\sigma.
\end{align}

Next, the cumulative distribution of the eigenvalues that do not fall within the aforementioned three ($\lambda_1, \lambda_2, \lambda_N$) will be calculated.
The cumulative distribution of the eigenvalues of the random matrix with a mean of zero, $\bar{J}$, follows Wigner's semicircle law~\cite{wigner1958semicircle} as follows:
\begin{align}
P_0(\lambda<x)=\int_{-2\sqrt{N}\sigma}^x \sqrt{4N\sigma^2-t^2} dt. \label{eq:wigner}
\end{align}
Based on the analysis presented in the formula \eqref{eq:q_eigen}, it can be posited that the eigenvalues of the coupling matrix $J$ also follow the semicircle law well, except for the largest eigenvalue $\lambda_1$.
Indeed, Tao \textit{et al.}\ have provided a proof of the semicircle law for the case of finite averages~\cite{tao2013semicircle}.
The eigenvalues $\lambda_i,\ (i=2,3,\dotsc,N)$ therefore follow Eq.~\eqref{eq:wigner} well.

Based on the above analysis, we can find the cumulative distribution of the eigenvalues of the coupling matrix $J$.
For simplicity, we normalize the eigenvalues as follows:
\begin{align}
\hat{\lambda} = \frac{\lambda-\lambda_N}{\lambda_1-\lambda_N}. \label{eq:eigen_normed}
\end{align}
Consequently, the expected value of the second largest eigenvalue, $\hat{\lambda}_2$, is approximated by the following expression by using Eqs.~\eqref{eq:eigen_1}--\eqref{eq:eigen_N}:
\begin{align}
r \equiv \frac{4\sqrt{N}\mu\sigma}{(\sqrt{N}\mu+\sigma)^2}. \label{eq:eigen_2_normed}
\end{align}
The remaining eigenvalues, $\hat{\lambda}_i\ (i=2,3,\dotsc,N)$, obey Wigner's semicircle law in Eq.~\eqref{eq:wigner} well.
Therefore, by appropriately determining the coefficients so that the function is continuous at $x=r$, we obtain the following:
\begin{align}
P\left( \hat{\lambda}<x \right)=
\begin{dcases}  
1+\frac{x-1}{(1-r)N} & \text{for $x>r$}, \\
\frac{1}{N}+\frac{4(N-2)}{\pi N r^2} \displaystyle\int_{0}^{x}\sqrt{r^2-(2t-r)^2}dt & \text{for $x\leq r$}.
\end{dcases}
\label{eq:eigen_dist_small_J}
\end{align}
This is an approximate function for the cumulative distribution of all eigenvalues when $\sqrt{N}\mu \gg \sigma$.
Although Eq.~\eqref{eq:eigen_dist_small_J} was derived in the perturbative regime, as we will see later, it is also a suitable approximation in the region where $\sqrt{N}\mu \approx \sigma$.

In what follows, we will consider the case where the variance of the coupling coefficients, $\sigma$, is greater than $\sqrt{N}\mu$.
In this case, the cumulative distribution of eigenvalues can be effectively approximated by Eq.~\eqref{eq:wigner}, as the region is close to a random matrix with a mean of zero.
Thus, we have
\begin{align}
P\left( \hat{\lambda}<x \right)=
\frac{1}{N} + \frac{4(N-1)}{\pi N} \displaystyle\int_{0}^{x}\sqrt{1-(2t-1)^2}dt.
\label{eq:eigen_dist_large_J}
\end{align}
From the above analysis, the approximation of the cumulative distribution of the eigenvalues of the coupling matrix in Eq.~\eqref{eq:q_gauss} is given by Eq.~\eqref{eq:eigen_dist_small_J} for $\sqrt{N}\mu \geq \sigma$, and by Eq.~\eqref{eq:eigen_dist_large_J} for $\sqrt{N}\mu < \sigma$.
The cumulative distribution of the eigenvalues obtained using these functions is referred to as the approximate cumulative distribution.

We derive the effective rank $K^*$ for determining the hyperparameters of FM using the approximate cumulative distribution.
We will determine the effective rank of the FM by ignoring eigenvalues whose ratio to the largest eigenvalue, $\hat{\lambda}_1=1$, is less than $0<\alpha<1$.
The determination of the threshold value based on the eigenvalue is a frequently employed method~\cite{cattell1966thescree,gavish2014theoptimal,jolliffe2002principal}.
The utilization of a threshold value based on the eigenvalues facilitates the determination of a reasonable rank of the FM.
The number of eigenvalues that are ignored can be estimated as $N\cdot P(\hat{\lambda}<\alpha)$ using Eqs.~\eqref{eq:eigen_dist_small_J} or \eqref{eq:eigen_dist_large_J}.
In consequence, the effective rank is given as follows:
\begin{align}
K^*(\alpha) =  N\left(1-P\left( \hat{\lambda}<\alpha \right)\right).
\label{eq:k_pred}
\end{align}
In the present paper, $K^*$ in Eq.~\eqref{eq:k_pred} is referred to as the predicted rank of FM.
The predicted rank, $K^*$, is obtained by performing the integral as follows.
When $r<1$, we have
\begin{align}
K^*(\alpha) =
\begin{dcases}  
\frac{1-\alpha}{1-r} & \text{for $\alpha>r$}, \\
1+\frac{N-2}{\pi} f\left(\frac{\alpha}{r}\right) & \text{for $\alpha\leq r$}.
\end{dcases}
\label{eq:k_pred_explicit_small_J}
\end{align}
Here, $f(x)=\arccos(2x-1)-2\sqrt{x(1-x)}(2x-1)$.
When $r=1$, the predicted rank is given as
\begin{align}
K^*(\alpha) = \frac{N-1}{\pi} f(\alpha).
\label{eq:k_pred_explicit_large_J}
\end{align}

\section{Numerical Results}
\label{sec:results}

This section presents the results of numerical experiments that demonstrates the advantage of the low-rank approximation method for initializing the FM described in Sec.~\ref{sec:init_with_eigen_decomposition} in reproducing the approximate coupling matrix, as compared to random initialization methods described in Sec.~\ref{sec:random_initialization_methods}.
Furthermore, we demonstrate that the FM initialized by the low-rank approximation can accurately represent the target model for the predicted rank $K^*$ of the FM, which is estimated using random matrix theory.
In these analyses, the Sherrington--Kirkpatrick (SK) model~\cite{kirkpatrick1978infinite,panchenko2013sk} is employed as the target approximate Ising model.
The Hamiltonian of the SK model is given in Eq.~\eqref{eq:apporx_ising_model}.
The constant term $c$ and the magnetic field coefficients $h_{i},\ (i=1, ..., N)$ are equal to zero, and the coupling coefficients $J_{ij}$ follow a Gaussian distribution:
\begin{align}\label{eq:sk_distribution_Jij}
J_{ij} \iid \mathcal{N}\left( \frac{J_{0}}{N}, \frac{J_{1}^{2}}{N} \right).
\end{align}
Here, $J_{0} \in \mathbb{R}$ and $J_{1} \in \mathbb{R}$ are parameters that represent the mean and variance, respectively, normalized with respect to the input dimension $N$.
In this paper, we assume that $J_{0}=1$.
To assess the precision of the approximation of the coupling matrix, we utilize the coupling error in Eq.~\eqref{eq:coupling_error}.

\subsection{Comparison of Initialization Methods}
We conducted numerical experiments to evaluate the performance of the FM initialization method.
An instance of the SK model was created by generating the coupling coefficients in accordance with Eq.~\eqref{eq:sk_distribution_Jij}.
Using the instance, we generated training data comprising spin configurations and the corresponding SK model energies.
Each data point in the training data is unique, i.e.\ the spin configurations are different from one another.
The FM was initialized using two distinct approaches: the low-rank approximation-based initialization and the random initialization, as detailed in Sec.~\ref{sec:initialization_methods}.
Subsequently, each FM was trained using the aforementioned training data.
Since the mean and variance of the coupling coefficient distribution are given by Eq.~\eqref{eq:sk_distribution_Jij}, the population mean and variance were used instead of the sample mean and unbiased variance for the purpose of random initialization in the numerical experiment.
The optimization algorithm used in the training is AdamW~\cite{loshchilov2018decoupled}.
The learning rate was set to $0.001$, and the number of learning epochs was set to $10{,}000$.
These parameters were selected based on the results of preliminary numerical experiments, which demonstrated that the training error converged sufficiently with these parameters.
The coupling error $\Delta J$ in Eq.~\eqref{eq:coupling_error} was calculated just after the initialization of the FM and every $100$ epochs during training, resulting $101$ measurements of the coupling error.
Then, the least coupling error in the $101$ measurements was seleced.
The above calculation was repeated $50$ times by changing training data to obtain a distribution of the coupling error.
In the case of the random initialization methods, we generated initial random parameters of the FM for each of $50$ calculations.

\begin{figure}[tbp]
  \centering
  \includegraphics[width=\textwidth]{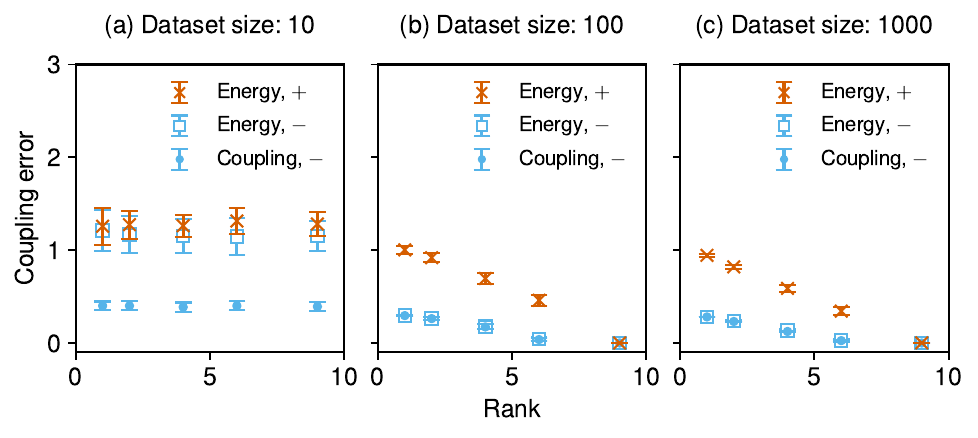}
  \caption{Coupling errors as a function of the rank of the FM model for random initialization methods. The number of training data is (a) $10$, (b) $100$, and (c) $1{,}000$. The coupling errors for the random initialization method based on energy with the positive FM are represented by red crosses, and the negative FM by open blue squares. The coupling errors for the random initialization method based on coupling coefficient distribution with the negative FM are shown by filled blue circles. The error bar represents standard deviation of the distribution.}
  \label{fig:coupling_error_among_random_inits}
\end{figure}

The numerical experiments demonstrated that, among the random initialization methods introduced in Sec.~\ref{sec:random_initialization_methods}, the method that initializes the negative FM based on the coupling coefficient distribution exhibited the greatest performance.
\figurename~\ref{fig:coupling_error_among_random_inits} illustrates the rank dependence of $\Delta J$ in Eq.~\eqref{eq:coupling_error} for each random initialization method.
The input dimension is $N=10$, and the variance parameter of the coupling coefficient of the SK model is $J_{1}=0.1$.
The results show that when the size of the training data is relatively limited in comparison to the number of possible spin configurations $2^N$, the method of initializing the negative FM based on the coupling coefficient distribution can result in reduced coupling errors.
As the size of the training data increases, the difference between the two random initialization methods for the negative FM diminishes.
In the case of $K=9$, where the FM can express any coupling matrix, every random initialization method achieves a low error with a sufficient amount of training data.
In evaluating the FM initialization method in FMQA, the following two points must be taken into account.
First, low-rank FMs are employed in preference to full-rank FMs, with the objective of reducing calculation times and enhancing optimization performance.
Second, it is difficult to train the FM with a substantial amount of training data whose size is comparable to the number of possible spin configurations when $N$ is large.
For this reason, we conclude that initialization of the negative FM based on the coupling coefficient distribution is the most effective among the random initialization methods in FMQA.
Actually, we confirmed that the superiority of the initialization method of the negative FM based on the coupling coefficient distribution was evident when the input dimension is increased to $N=50$.
As the variance parameter is increased to $J_{1}=10$, the difference between the random initialization methods is diminished. 
Nevertheless, it is confirmed that the initialization of the negative FM based on the coupling coefficient distribution results in coupling errors that are comparable to those of the other random initialization methods.
Although \figurename~\ref{fig:coupling_error_among_random_inits} depicts the results for a single instance of the SK model,  similar results have been confirmed to be obtained for other instances.
Hence, among the random initialization methods, it was determined that the approach of initializing the negative FM based on the coupling coefficient distribution is the most effective and, robust to changes in the instance of the SK model.

\begin{figure}[htbp]
  \centering
  \includegraphics[width=\textwidth]{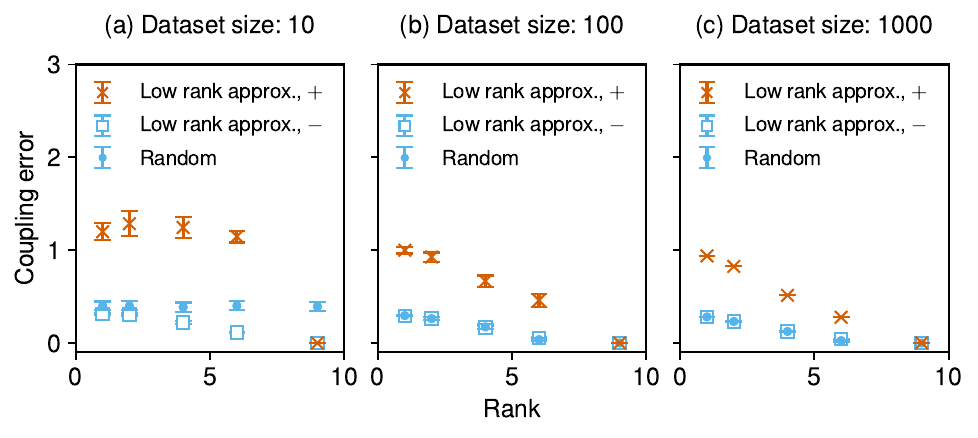}
  \caption{Coupling errors as a function of the rank of the FM model for the coupling-based random initialization with the negative FM and the initialization using the low-rank approximation. The number of training data is (a) $10$, (b) $100$, and (c) $1{,}000$. The coupling errors for the initialization method based on the low-rank approximation with the positive FM are shown by red crosses, and those with the negative FM are by open blue squares. The coupling errors for the random initialization method are dipicted by filled blue circles. The error bar represents standard deviation of the distribution.}
  \label{fig:coupling_error_random_init_vs_eigen_decomposition}
\end{figure}

Subsequently, we show that the initialization method using the low-rank approximation, as described in Sec.~\ref{sec:init_with_eigen_decomposition}, is more effective than the random initialization methods.
As a random initialization method, we employed the negative FM initialization method based on the distribution of coupling coefficient, since the method has been demonstrated to achieve the lowest coupling error within the random initialization methods, as previously described.
The results of comparing the coupling errors for these initialization methods, conducted under the identical experimental conditions as those employed in the previously described numerical experiment, are presented in \figurename~\ref{fig:coupling_error_random_init_vs_eigen_decomposition}.
The results demonstrate that initialization methods using the low-rank approximation can achieve reduced coupling errors when the FM is trained on relatively modest datasets.
Although the results presented here are for $N = 10$, the superiority of the initialization method using the low-rank approximation for $N = 50$ has also been confirmed.
Hence, the initialization method using on the low-rank approximation is more effective than the random initialization methods in reproducing the coupling matrix of the SK model.

\begin{figure}[htbp]
  \centering
  \includegraphics[width=0.5\columnwidth]{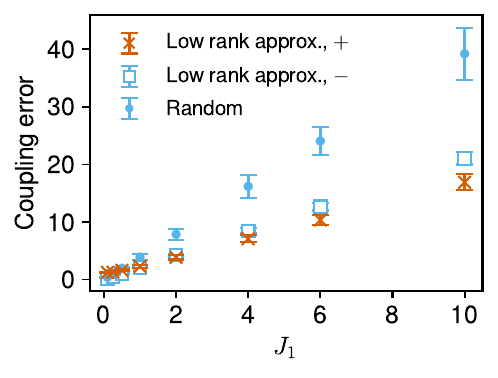}
  \caption{Coupling errors as a function of $J_{1}$ for the coupling-based random initialization with the negative FM and the initialization methods using the low-rank approximation. The number of training data is (a) $10$, (b) $100$, and (c) $1{,}000$. The coupling errors for the initialization method based on the low-rank approximation for the positive FM are represented by red crosses, and those for the negative FM by open blue squares. The coupling errors for the random initialization method are shown by filled blue circles. The error bar represents standard deviation of the distribution.}
  \label{fig:coupling_error_random_init_vs_eigen_decomposition_various_j}
\end{figure}

In comparison to a random initialization method, the FM initialization method using the low-rank approximation becomes increasingly advantageous as the variance $J_{1}$ increases.
For the random initialization method, we used the negative FM initialization method based on the dictribution of coupling coefficent.
\figurename~\ref{fig:coupling_error_random_init_vs_eigen_decomposition_various_j} illustrates the correlation between the coupling error and the value of $J_{1}$ for each initialization method.
The input dimension is $N=10$, the rank of the FM is $K=4$, and the dataset size is $10$.
The coupling error increases with $J_{1}$.
This is due to the fact that the absolute value of the coupling coefficient increases in direct proportion to $J_{1}$.
As the results show, the difference in coupling errors between the two initialization methods is amplified as the variance $J_{1}$ increases.
This behavior was the same even when the input dimension is $N=50$.
It can thus be concluded that the initialization method for the FM using the low-rank approximation is superior even when the value of the variance, $J_{1}$, is altered.

\subsection{Analysis with Random Matrix Theory}
In order to assess the validity of the approximate cumulative distributions in Eqs.~\eqref{eq:eigen_dist_small_J} and \eqref{eq:eigen_dist_large_J} derived through random matrix theory, the following numerical experiments were conducted.
Since the coupling coefficients of the SK model are distributed according to a Gaussian distribution, the analysis presented in Sec.~\ref{sec:approximation_error_analysis_random_matrix_theory} can be applied.
The coupling matrix of the SK model in Eq.~\eqref{eq:sk_distribution_Jij}, in a precise sense, does not correspond to a Gaussian orthogonal random matrix due to the presence of diagonal elements.
However, the diagonal elements can be eliminated because that operation does not change the Hamiltonian.
In this study, we address the task of estimating the appropriate hyperparameter $K$ for training FM using training data sampled from the Hamiltonian of the SK model.
Here, we fixed $J_0=1$, and generated the coupling matrix using Eq.~\eqref{eq:q_gauss} for $J_{1}=0.03$, $0.1$, $0.3$, and $1.0$ with system sizes of $N=10$, $50$, and $100$.
We generated $20$ instances for each combination of $J_1$ and $N$.
The eigenvalues were calculated using the eigenvalue decomposition, and the cumulative probability distribution was plotted after normalization using Eq.~\eqref{eq:eigen_normed}.

\begin{figure}[htbp]
    \centering
    \includegraphics[width=\linewidth]{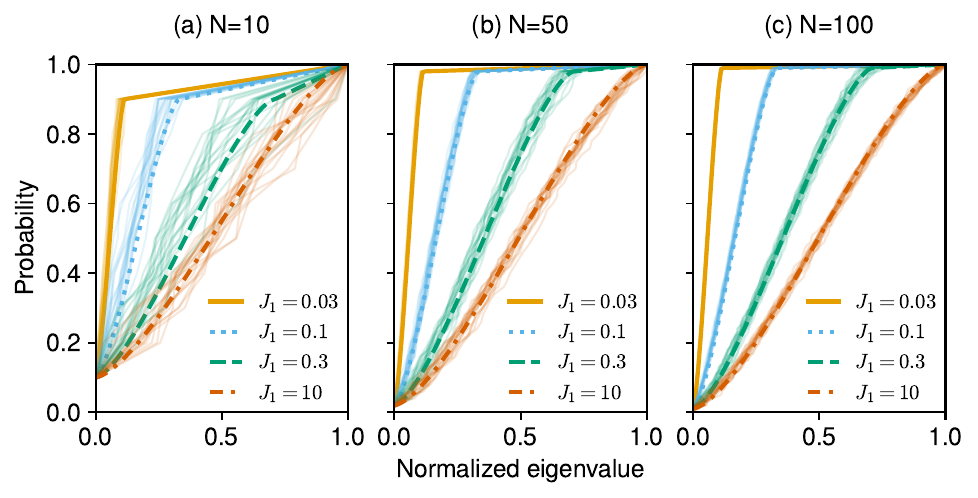}
    \caption{Numerical and theoritical cumulative probability distributions. The system size is (a) $10$, (b) $50$, and (c) $100$. The thin lines represent numerical cumulative probability distribution. The thick lines are results obtained by the analysis with random matrix theory.}
    \label{fig:cumulative_dist_with_predicted_dist}
\end{figure}

The result of the approximate cumulative distribution in addition to the cumulative distribution of the normalized eigenvalues is shown in \figurename~\ref{fig:cumulative_dist_with_predicted_dist}.
The thick lines represent the approximate cumulative distributions based on random matrix theory.
The other thin lines represent plots of cumulative distribution of the normalized eigenvalues obtained for each pair of $J_1$ and $N$.
As can be observed, the experimentally determined cumulative distribution aligns with the theoretical lines of the approximate cumulative distribution.
Additionally, it was observed that as the input dimension $N$ increases the agreement between the theoretical and experimental values improves.
This is due to the fact that the variance of the coupling matrix~\eqref{eq:q_gauss} of the SK model converges to zero.
For the case of $J_{1}>J_0=1$, the results were almost the same as those for $J_1=10$ because the random matrix with a sufficiently large variance compared to the mean behaves like a random matrix with a mean of zero.
In fact, as the value of $J_1$ increases, the singularity of the second largest eigenvalue exhibits a positive increase.
At $J_1=10$, the singularity is no longer discernible.
The above shows that the approximate cumulative distribution proposed in this study is largely independent of the specific instance details and can accurately describe the cumulative distribution of eigenvalues of the coupling matrix of the SK model with high accuracy.

We show a relationship between the measured coupling error and the predicted rank $K^*$ given by the Eqs.~\eqref{eq:k_pred_explicit_small_J} or \eqref{eq:k_pred_explicit_large_J}.
Let us first consider the case $J_{1}=0.1$.
Approximation is made to ignore eigenvalues below a fraction $\alpha=0.15$ of the normalized largest eigenvalue.
In this case, Eq.~\eqref{eq:k_pred_explicit_large_J} produces $K^*=5.47$ for $N=10$ and $K^*=27.82$ for $N=50$.
Similarly, in the case of $J_{1}=10$, the predicted rank can be calculated by setting the ratio $\alpha$ appropriately.
In order to achieve an error equal to that of $J_{1}=0.1$, you can obviously use the same value as the predicted rank $K^*$ mentioned above.
This is because eigenvalues other than the largest eigenvalue are almost unaffected by the modulation introduced by $J_{1}$, and follow Wigner's semicircle law well.
In other words, the value of $K^*$ is found to be $K^*=5.47$ for $N=10$ and $K^*=27.82$ for $N=50$ without determining the ratio $\alpha$ explicitly.

\begin{figure}
    \centering
    \includegraphics[width=0.48\linewidth]{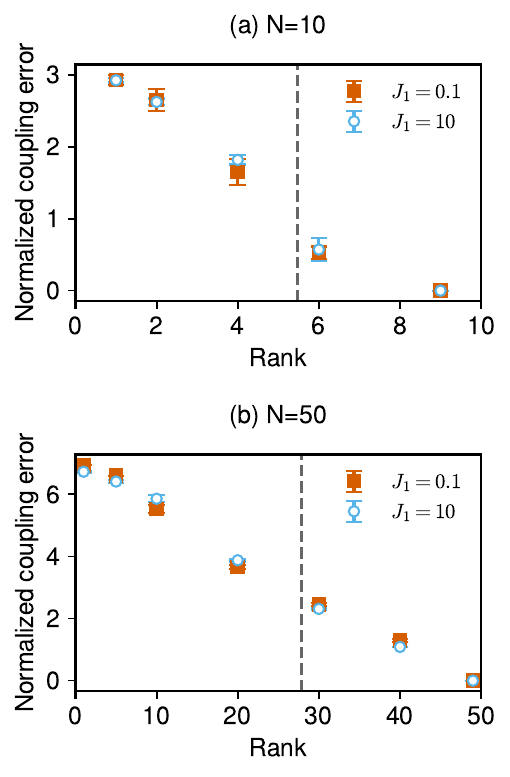}
    \caption{Coupling error and predicted rank. The system size is (a) $10$ and (b) $50$. Coupling errors obatained by numerical calculations as a function of the rank of the FM are represented by filled red squares for $J_1 = 0.1$ and open blue circles for $J_1 = 10$. The dashed black vertical line shows the predicted rank $K^*$.}
  \label{fig:error_with_predicted_rank}
\end{figure}

\figurename~\ref{fig:error_with_predicted_rank} shows the plot of the coupling error of the negative FM obtained by using the low-rank approximation together with the above results.
The predicted rank $K^*$ is indicated by the dashed black vertical lines.
Here, the coupling error was normalized by dividing by the variance $J_{1}$.
At the predicted rank, the normalized error is largely suppressed compared to the case of lower ranks.
Smaller values of $\alpha$ would shift the predicted rank to the right, yielding a model with a more suppressed coupling error.
As illustrated in \figurename~\ref{fig:error_with_predicted_rank}, the coupling error demonstrates a consistent behavior independent of $J_1$, indicating that the predicted rank is a good predictor of model performance.

\section{Conclusion}
\label{sec:conclusion}
In this paper, we present a novel initialization method for accurately reproducing a coupling matrix using the FM when the coupling matrix of an Ising model is provided approximately.
The results of numerical experiments demonstrate that the initialization method based on the low-rank approximation is more effective than the random initialization methods in obtaining an approximate coupling matrix with greater accuracy.
Furthermore, the approximation performance of the low-rank initialization method was evaluated using random matrix theory.
The findings indicated that as the input dimension increased, the approximation performance of the low-rank initialization method exhibited diminished dependency on the specific problem instance.

It has been demonstrated that, if it is known that the elements of the true coupling matrix follow a Gaussian distribution, the predicted rank $K^*$ can be analytically obtained using the equations in either Eq.~\eqref{eq:k_pred_explicit_small_J} or Eq.~\eqref{eq:k_pred_explicit_large_J}.
The application of these methods has the potential to significantly reduce the computational complexity of the algorithm in Sec.~\ref{sec:init_with_eigen_decomposition}.
When the coupling matrix of the true model is known, it is possible to derive the requisite rank to keep the error given by Eq.~\eqref{eq:theoretical_coupling_error_low_rank_approx} below an allowable value, using eigenvalue decomposition.
However, since the eigenvalue decomposition generally has a computational complexity of $O(N^3)$, calculation of $K^*$ through the eigenvalue decomposition requires longer computational time as the input dimension increases.
On the other hand, if the normality of the coupling matrix can be confirmed using a test such as the Kolmogorov--Smirnov test~\cite{sheskin2003ks}, it is possible to estimate the predicted rank using Eqs.~\eqref{eq:k_pred_explicit_small_J} or~\eqref{eq:k_pred_explicit_large_J}.
In particular, if the mean and variance of the coupling matrix are known, the predicted rank $K^*$ can be estimated with a computational complexity of $O(1)$.
In this case, the truncated eigenvalue decomposition up to the top $K^*$ positions~\cite{hansen1999tsvd} provides a reduction in computational cost to $O(K^*N^2)$.
From the above, this method also has the effect of enhancing the method presented in Sec.~\ref{sec:init_with_eigen_decomposition}.
However, the verification of this assertion will be a subject of future research.

The findings of this study are beneficial for those engaged in optimization using FMQA.
One potential application of FMQA is to enhance the approximation accuracy of problems for which the coupling matrix of Ising model is provided approximately.
In this particular application, it is essential to construct an FM that accurately reproduces the approximate coupling matrix.
In light of the warm-start method, it is anticipated that a more precise FM model can be constructed by identifying the initial parameters of the FM from the approximate coupling matrix.
The findings of this study demonstrate that the initialization method for the FM based on the low-rank approximation allows for the accurate expression of the approximate coupling matrix.
It is expected that future research will yield the development of FMQA with high performance based on the findings of this study. In particular, elucidating the performance of FMQA that utilizes the warm-start method represents an intriguing avenue for future investigation.

\section*{Acknowledgments}
The authors would like to thank Kotaro Tanahashi for useful discussions.
This work was partially supported by JSPS KAKENHI (Grant Number JP23H05447), the Council for Science, Technology, and Innovation (CSTI) through the Cross-ministerial Strategic Innovation Promotion Program (SIP), ``Promoting the application of advanced quantum technology platforms to social issues'' (Funding agency: QST), JST (Grant Number JPMJPF2221). This paper is partially based on results obtained from a project, JPNP23003, commissioned by the New Energy and Industrial Technology Development Organization (NEDO).
The authors wish to express their gratitude to the World Premier International Research Center Initiative (WPI), MEXT, Japan, for their support of the Human Biology-Microbiome-Quantum Research Center (Bio2Q).

\bibliography{references}

\end{document}